# SemTK: A Semantics Toolkit for User-friendly SPARQL Generation and Semantic Data Management

Paul Cuddihy[1], Justin McHugh[2], Jenny Weisenberg Williams[1],
Varish Mulwad[1] and Kareem Aggour[1]

[1] GE Global Research, Niskayuna NY 12309, USA
[2] Kobai, Pleasanton, CA 94588, USA
`cuddihy@ge.com,justin@kobai.io`
`{weisenje,varish.mulwad,aggour}@ge.com`

**Abstract.** While the use of knowledge graphs has exploded in recent years, there exist few tools and mechanisms for users to explore, query and manage semantic data in knowledge graphs. A lack of user-friendly tools to construct SPARQL queries has been a barrier preventing the wide adoption of Semantic Web technologies by domain experts and application developers alike. There are a variety of tools and techniques to convert CSV and relational data to RDF, but to the best of our knowledge there is no integrated, user-friendly toolkit for data triplification and ingestion into a triple store. In this paper, we present the Semantics Toolkit (SemTK), an open source project that allows user-friendly querying and semantic data management. Through its user interface, SemTK allows users to convert CSV data into RDF triples and ingest them into a triple store. It also allows users to visually explore the ontology and construct SPARQL queries via a drag-and-drop interface. SemTK also provides novel SQL stored procedure-like support for saving and executing semantic queries with run-time constraints. Additionally, SemTK provides REST APIs for all of its functionality including allowing data ingestion and queries to be executed programmatically, dramatically simplifying the deployment of knowledge-driven applications. The Semantics Toolkit is open-sourced under the Apache License, Version 2.0 and is available at https://github.com/ge-semtk/semtk.

## 1 Introduction

With the major success of several commercial artificial intelligence and cognitive applications such as Siri, Cortana, Google Now and Watson, knowledge graphs have been rapidly gaining traction in industry. However, the Semantic Web technology stack that provides a foundation to construct, query and maintain knowledge graphs poses a significant barrier to their adoption by both non-semantic subject matter experts in scientific and industrial communities, as well as by application developers building knowledge graph-driven applications. A breakout session on "Broadening the Base" at the 2018 U.S. Semantic Technologies Symposium[1] echoed these views.

Tools such as Protégé [1] and SADL[2] [2] have made rapid strides in reducing these barriers for ontology design and creation. However, there exist very few tools with the same level of maturity to explore, query and manage semantic data in knowledge graphs. At GE we experienced this problem—while our industrial domain experts can design and develop ontologies using SADL, the lack of user-friendly tools for data triplication and querying was a major hindrance to the broader adoption of Semantic Web technologies. As we set out to build the GE Turbine Engineering Data System [3], an ontology-based data access system for navigating equipment test data, it was clear that success depended upon the ability of engineers to write their own queries. We also saw bottlenecks in the ability to store and execute queries, and perform data ingestion. These needs were fulfilled in the form of a general-use Semantics Toolkit: SemTK.

The Semantics Toolkit (SemTK) is designed to lower the barrier to using semantic technologies by making semantics accessible in a user-friendly manner to both subject matter experts and application developers. With an "ontology first" approach, SemTK has been designed in the context of the needs of a large industrial business, and in the context of the migration of large disparate data sources into Linked Data. Through its SPARQLgraph interface, SemTK allows both domain experts and application developers to upload ontologies, browse existing ontologies, upload data with the help of a drag-and-drop CSV-to-triples mapping generator and data ingestor, and finally a visual, graphical drag-and-drop approach for SPARQL query generation to access and explore both the ontology and the data. Further, SemTK allows application developers to save the data triplification mappings and graphically construct and save queries and access them programmatically via REST APIs, dramatically reducing knowledge-driven application development time. To the best of our knowledge, SemTK is the first open source system that provides a seamlessly integrated experience to perform all of these tasks in a single toolkit.

The paper is organized as follows: Section 2 details the key features and capabilities of the Semantics Toolkit and its SPARQLgraph user interface. Section 3 discusses the availability and reusability of the system, and Section 4 describes the related work. Section 5 concludes the paper and outlines future work.

## 2    The Semantics Toolkit

SemTK is comprised of a suite of Java REST services that work with SPARQL 1.1 compliant triple stores. On top of those services is a web-based tool called SPARQLgraph, which is the primary interface through which models are explored and provides drag-and-drop interfaces for query generation and data importing. Therefore, this interface can be used to demonstrate the wide variety of SemTK features that lower the barriers to these tasks so that semantic scientists can work more efficiently, and application developers can quickly ingest and query data.

---

[1] http://us2ts.org/posts/program/

[2] http://sadl.sourceforge.net/

### 2.1 Basic Concepts

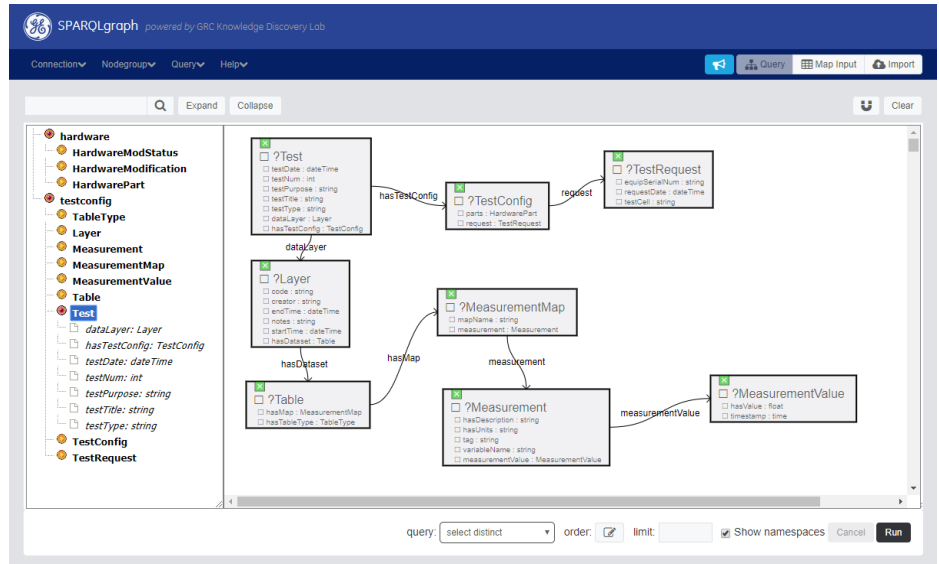

**Fig. 1.** The SPARQLgraph screen capture illustrates three basic building blocks of all SemTK operations: (i) the connection, (ii) the ontology information cache (oInfo), and (iii) the nodegroup. This simplified example ontology represents a collection of equipment test data, with results organized into layers of measurements.

**Connections.** A SemTK connection consists of the URL and graph names of the SPARQL endpoints that store the ontology and the data. SemTK allows the ontology and the data to be stored at different endpoints. Additionally, the "model domain" is a regex used to select the URI's of classes and attributes contained in an ontology. Entities referenced by the ontology that do not fall within the ontology domain are often primitives such as XMLSchema numbers, strings, and dates. The model domain typically maps back to part of/complete base URI of the ontology. Once a connection is defined, users can drag and drop an OWL ontology file via the Import tab in the SPARQLgraph interface to load the ontology in the given connection.

**Ontology Information (oInfo).** Users first select a desired connection to read existing OWL ontologies. Ontology information is then loaded via SPARQL queries against the endpoints specified in the connection into a local cache for fast access. This cache is called "oInfo", and includes classes, subclass relationships, class properties and types, and permitted values for enumeration classes. This information is displayed to the user as a hierarchical list in the ontology panel, the left-hand pane of SPARQLgraph, as shown in Fig. 1. This panel allows users to explore the loaded domain ontologies in preparation for querying or data ingestion. The hierarchical list conveys the subclass relationship between classes. On the expansion of each class, it also displays the associated datatype and object properties along with their respective ranges. To deal with the challenge of ontologies with 100s to 1,000s of classes and properties, this panel provides a search control enabling users to quickly drill down and highlight any part of the ontology matching the search string.

**Nodegroup.** All SemTK querying and ingestion is based on a data structure called a "nodegroup" shown on the right-hand pane of Fig. 1. A nodegroup is a powerful representation crucial to almost every SemTK function. Each node in the nodegroup represents a class variable in a query, and has a unique name and a list of properties. The properties are split such that those outside the ontology domain are listed first. These properties are often primitive OWL datatypes, but may also be properties linked to objects outside the domain (e.g., objects in DBpedia). All properties within the connection domain are displayed, regardless if they are being used in a query or not. Users construct nodegroups by dragging classes from the oInfo panel.

## 2.2    Query Generation

With the ontology information loaded and searchable, the user can begin drag-and-drop query generation and query execution against the data endpoints. SemTK currently supports the generation of the following subset of SPARQL query types available in SPARQL 1.1: select, construct, ask, delete and insert. Query generation with SemTK allows for features including regex, values clauses, counting, limits and optional values.

**Pathfinding.** In a large ontology, queries commonly incorporate entities separated by many links. This requires a complex SPARQL query consisting of many clauses representing the chain of relationships between the two entities. Further, there may be more than one possible path to connect the two entities. SPARQLgraph and SemTK simplifies the construction of SPARQL queries over such ontologies with a feature known as pathfinding. Pathfinding greatly enhances the ease-of-use in building a nodegroup; as a new class is dragged from the ontology panel onto the nodegroup canvas, SemTK uses a slight variation of the well-known A* algorithm[3] to suggest various paths by which the new class might connect to the existing nodes in the nodegroup. Pathfinding allows users to drop two classes on the canvas and select the desired connecting path. SemTK uses object property ranges to identify possible classes in a path.

Pathfinding is driven by an implementation of the A* algorithm modified with stopping logic. The search is limited to a maximum path length (typically 10 links) and/or search time (typically 30 seconds). Further, search path lengths can be limited such that once a path of length n is found, searching ends when all paths of length (n+m) have been explored. This search, restricted to local paths instead of a theoretically complete set, provides an indispensable feature for the efficient assembly of nodegroups. The combination of pathfinding with the performance enabled by the ontology cache allows users to quickly build complex nodegroups, and subsequently auto-generate SPARQL.

---

[3] https://en.wikipedia.org/wiki/A*_search_algorithm.

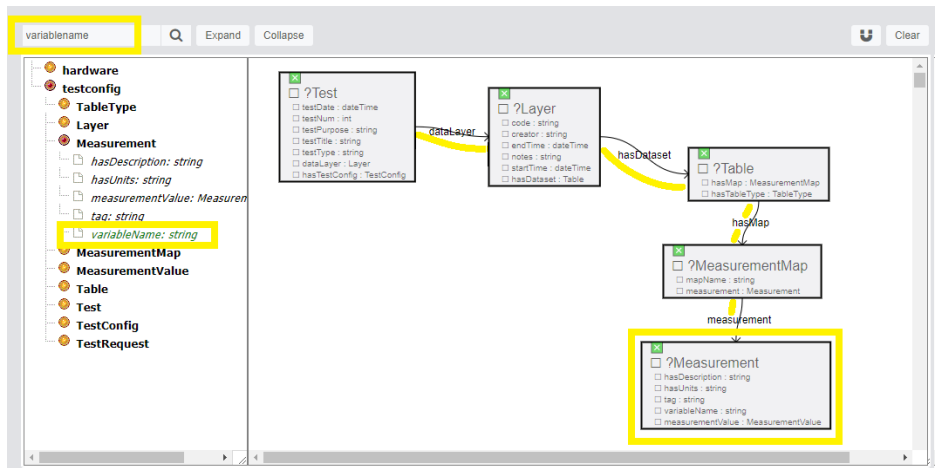

**Fig. 2.** Pathfinding example: In this example, a user has started with a *?Test* node, and searched to find *variableName* properties. After finding this in the *?Measurement* class, the user dragged that class onto the canvas. The yellow highlighting shows the results of automatic pathfinding. Had there been multiple short paths to *?Measurement*, the user would have been presented with an options dialog to choose the intended path.

**Generating SPARQL.** Once a nodegroup has been constructed, SemTK can automatically generate SPARQL to select data. Consider the case where a user has clicked to select *testNum* from *?Test* and *variableName* from *?Measurement* to be returned by a select query. SemTK walks the nodegroup and generates clauses for each node, including clauses for class type, relationships to other nodes, and relationships to properties. Table 1 shows the resulting SPARQL.

**Adding Constraints to the SPAQRL query.** Users can add constraints via FILTER and VALUES clauses by clicking on any property in the nodegroup. Users can either manually hand-edit the clauses or can leverage SemTK's *Automated VALUES Generation*. SemTK leverages its automatic query generation to provide a powerful ability to suggest all valid values for a property based on the current nodegroup and the contents of the data endpoints. SemTK converts the current nodegroup into SPAQRL, queries under the hood for all possible values, and then presents them in list form to the user. The queries used to populate the list are SELECT DISTINCT queries with two modifications: only the target variable is returned, and all FILTER or VALUES clauses are removed for that variable. Only those values that satisfy the constraints and relations specified in the nodegroup are returned. In the *?Test / variableName* example, when the user attempts to build a VALUES clause on *variableName*, SemTK can execute the query similar to the SELECT in Table 1. Only *?testNum* needs to be dropped from the SELECT clause to find all available *variableNames* attached to *?Test* with the path shown in the nodegroup. A search control helps users search through the list of values, and a VALUES clause is automatically generated with the user's selection(s).

**Table 1.** Auto-generated select query.

```
prefix XMLSchema:<http://www.w3.org/2001/XMLSchema#>
prefix testconfig:<http://com.ge.research/sample/testconfig#>
select distinct ?testNum ?variableName
     FROM <http://iswc/data>
     FROM <http://iswc/model>
 where {
   ?Test a testconfig:Test .
   ?Test testconfig:testNum ?testNum .
   ?Test testconfig:dataLayer ?Layer .
     ?Layer testconfig:hasDataset ?Table .
       ?Table testconfig:hasMap ?MeasurementMap .

         ?MeasurementMap testconfig:measurement ?Measurement .
           ?Measurement testconfig:variableName ?variableName .
}
```

This functionality becomes more powerful when it is chained together. For example, if the user were to limit *variableName* to "Temperature" then ask for possible values for *?Test*'s *testNum*, the same process would result in a list of only *testNum* values that are attached to *?Test* instances which have the given path to *?Measurements* with *variableName* "Temperature." Using this iterative method of building VALUES clauses, a user can quickly down-select and navigate through complex data.

**Runtime Constraints.** An important SemTK variation on FILTER and VALUES clauses is a runtime constraint capability. If a property is flagged as runtime constrainable, it's value is meant to be provided when the query is executed, mimicking functionally commonly associated with stored procedures in the relational database world. Using the automated VALUES generation feature, a user can easily be offered a list of likely values for the runtime constraints.

**Optionals.** When constructing a query, a user may mark a class or property as optional, indicating that they wish to retrieve the requested data whether the marked class or property is present or not. In this case, SemTK query generation encloses all nodes downstream (or upstream) of the connecting arrow inside a SPARQL OPTIONAL clause. Clauses are nested as the SPARQL is generated.

**Additional Clauses.** The user interface enables the specification of a maximum number of results to return, which SemTK implements by adding a LIMIT clause to the query. It allows sorting on all SPARQL IDs being selected. Further, users can select the "count" query type to return the number of results rather than the result contents. In this case, SemTK transforms the original select query into a count query by wrapping it using "SELECT (COUNT(*) as ?count)".

**Delete Queries.** A user may construct a nodegroup representing data to be deleted from the triple store. SemTK generates DELETE queries from such nodegroups. Deletion is straightforward for class attributes (including links between nodes), and consists of simply removing any instance of the given attribute from instances of the given class, taking into account any surrounding constraints. In contrast, in the case of deleting instances of classes, the user may choose from several modes to control the scope of the deletion. The simplest mode removes triples of the specified class. Other modes provide the option to delete all triples with the node variable in the subject or object, or to limit the deletion to predicates in the loaded model, or predicates in the nodegroup. Together, these DELETE modes provide the user with a wide range of options to support practical applications.

**Query Optimization.** SemTK's SPARQL query generation required proactive optimization to maintain acceptable performance. In early testing, the benchmark triple stores (OpenLink Virtuoso 7, Jena TDB[4] and Fuseki[5]) showed poor performance when attempting to execute generated queries. These early queries were generated by naïvely stepping through the contents of the nodegroup without regard to the impact of clause ordering on the SPARQL query performance.

To optimize query performance, SemTK orders clauses from those expected to be most specific to least specific. The system assumes that the relationships described in the ontology are directional. Any relationship not stated explicitly as being bi-directional is assumed to be outgoing from the subject to the object. It is also assumed that requested instances of classes with no incoming relationships are, in many cases, more tightly specified than other instances. This is because, in the sort of use cases for which the nodegroup is most useful, these instances often have outgoing connections. These instances are placed first in the generated queries, followed immediately by their outgoing connections. Applying these outgoing connections act as constraints. This ordering has the effect of decreasing the search space the SPARQL engine must examine to fulfill the later clauses. By the time the least specified instances are bound, the potential space has been decreased because the outgoing connections from the more specified entries have limited their scope.

The use of the above technique produced significant improvements over the original naïve implementation. In the case of Virtuoso, this led to greatly improved query execution times, making SPARQLgraph usable as a practical, interactive web-based tool. In the case of Fuseki and Jena TDB, it resulted in rapid responses instead of queries which ran for multiple minutes before failing due to timeout errors.

### 2.3    Data Ingestion

SemTK also provides capabilities to convert CSV data into RDF triples and ingest them into a triple store. Using the same ontology-based interaction model as the rest of the system, data triplification and ingestion is intuitive and provides advanced features for checking the incoming data.

---

[4] https://jena.apache.org/documentation/tdb/
[5] https://jena.apache.org/documentation/serving_data/

The triplification and ingestion of data in SemTK takes place in three steps. The first step is the constructing a nodegroup that will be used to define the structure of the triples to be generated. The second is the drag-and-drop mapping of columns from a CSV (or table) to the constructed nodegroup. Third, the ingestion process itself is kicked off, applying the mapping to the input data.

**Ingestion Data.** Consider the example case of ingesting data against the hardware test ontology (**Fig.** 3). This data spans six classes of objects that must be linked together.

| test_number | layer_code | meas_tag | meas_name | meas_units | timestamp | value |
|---|---|---|---|---|---|---|
| 4242 | layer1 | temp | temperature | F | 2017-03-23T10:00:00 | 200.5 |
| 4242 | layer1 | temp | temperature | F | 2017-03-23T10:03:16 | 200.8 |
| 4242 | layer1 | temp | temperature | F | 2017-03-23T10:03:17 | 200.9 |
| 4242 | layer1 | st1 | status1 | % | 2017-03-23T10:00:00 | 0.01 |
| 4242 | layer1 | st1 | status1 | % | 2017-03-23T10:03:16 | 0.05 |
| 4242 | layer1 | st1 | status1 | % | 2017-03-23T10:03:17 | 0.72 |

**Fig. 3.** Sample CSV data to be ingested in the triple store

**Ingestion Nodegroup.** The structure of an ingestion nodegroup will look the same as one created for SELECT queries, except that many clauses not related to the structure of the nodegroup (FILTER, VALUES, OPTIONAL, SORT etc.) will be ignored. For the data in **Fig.** 3, the nodegroup shown earlier needs to be expanded to include the *Measurement* class.

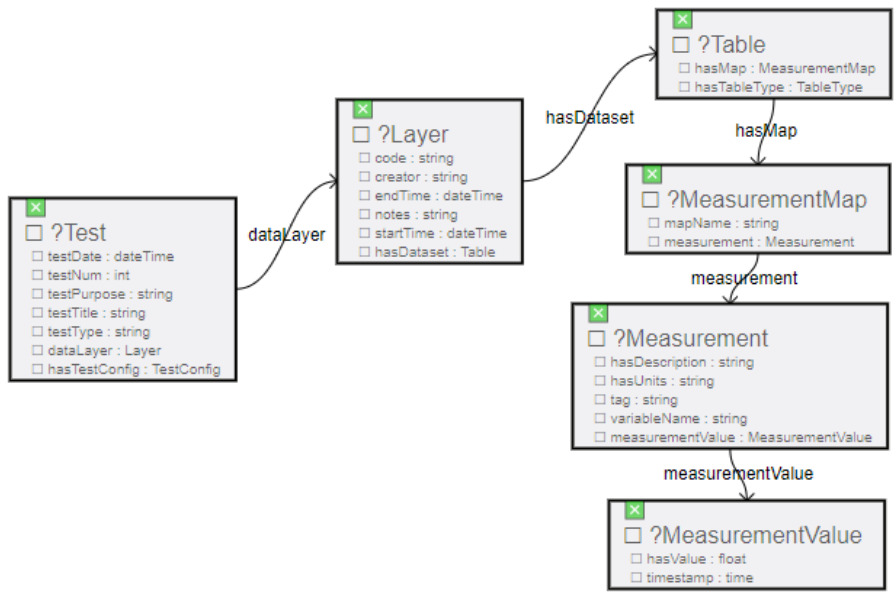

**Fig. 4.** User constructed node group for ingestion

**Ingestion Mapping.** SPARQLgraph provides drag-and-drop generation of ingestion mappings. The mappings associate one or more columns in the CSV data (**Fig.** 5 shown in green at right) to one or more attributes in the nodegroup (**Fig.** 5 shown on the left).

**Fig. 5.** The mapping tab allows a simple interface to associate CSV headers with the appropriate classes and properties

Additionally, transformations are available, allowing the column values to be altered before being assigned. Free text values can be added to the mapping as well.

When ingestion occurs, one copy of the nodegroup will be generated for each line of input data. The data is type-checked based on the ontology and then added to the nodegroup. Finally, the SPARQL-generation engine builds an INSERT statement and executes it against the data connection.

The example in Fig. 5 demonstrates two different strategies for generating URIs for the INSERT statements. The instance of *Test* has it's URI constructed from the text "TEST_" and the "TEST_NUMBER" data field. The triples for giving the URI a type and giving it a *testNum* property are executed for each row. In SPARQL 1.1, if these triples already exist, the insertion has no additional effect.

The *Measurement* is being inserted using a different and powerful strategy which is indicated by the darkened buttons. The user indicated that *Measurement* should be looked up based on the properties *hasUnits*, *tag*, and *variableName*. If a URI is found with all three properties matching the input data, that URI will be used. If a URI is not found, the ingestion can be instructed either to generate an error, or to create the missing object and assign it the appropriate property values. URI lookups are performed in a separate first pass of the ingestion process, and dramatically simplifies adding and linking new data to an existing knowledge base.

**Blank Values, Pruning, and GUIDs.** Mapping items that have no input columns associated with them are ignored. Blanks in the input data are also ignored. After a nodegroup is populated with input data, but before the INSERT query is generated, the nodegroup is pruned such that any leaf node with no data is removed. Occasional "empty" objects will remain when needed to serve as intermediate steps in a path between classes across the ingested data. These are assigned random UUIDs for their URIs.

## 2.4    SemTK Services: Application Development

While the SPARQLgraph tool is used extensively for uploading and exploring ontologies, creating nodegroups and ingestion templates, and exercising various other SemTK features, the powerful reuse of SemTK is often based on the microservices layer. Through SemTK, nodegroups can be built by domain experts without deep semantic expertise and application developers can execute them through REST calls. This provides complete separation between the software developers and domain experts when needed. As data is moved or ontologies are modified, new nodegroups can be added to an application. However, in practice we have found the drag-and-drop tools to be so user-friendly that many programmers with little familiarity with the Semantic Web can build and maintain their own nodegroups. The following are a few of the key features which have sped adoption of the SemTK service layer.

**Nodegroup Store.** SemTK tools typically convert nodegroups to JSON. The service layer is also able to store nodegroups in the triple-store and retrieve them by name. This allows an application to execute a query by name only. Similarly, data can be ingested by providing (i) the name of a nodegroup containing an import mapping and (ii) a data file.

**Asynchronous Job Handling.** The SemTK service layer include status and results services for asynchronous execution of longer jobs. The status service keeps an application up-to-date on the percentage of completion of a job, as well as failure messages when appropriate. Upon successful completion of a job, results are retrieved through a results service.

**External Data Connection.** While the triple store is an effective storage mechanism for many datasets, it is not suitable for many types of data, particularly those binary in nature (e.g., image data, files) or those requiring high overhead to store in a triple store (e.g. time series data). To address this, SemTK includes extensive capabilities for Ontology-Based Data Access (OBDA), enabling data stored outside of the triple store to be linked, browsed, and queried in domain-relevant terms as if it were part of the triple store. This functionality is achieved by storing metadata in the triple store that describes data stored elsewhere. SemTK's OBDA features are referred to as External Data Connection, and are closed-source at present. While this feature is not a focus of this paper, further discussion of this capability is described in [3].

# 3 Availability and Reusability

## 3.1 Reusability

Several business applications have been built using SemTK, which has proven to be generic and highly flexible to successfully support applications diverse in their data content, modeling approaches, and target environments. We demonstrate SemTK's reusability by describing two such applications.

**Turbine Engineering Data System.** The primary application that led to the development of SemTK is the Turbine Engineering Data System (TEDS) built for GE Power [3]. GE Power's engineering division performs large scale testing of turbines and turbine components, capturing extensive test configuration data (100's of parameters per test) and large amounts of measurement data (1Hz data collected for 10,000+ sensors for hours to months per test). TEDS, built on top of SemTK, makes this time series sensor Big Data accessible to design engineers via an ontology-driven user interface called SPARQLform (a simplified version of SPARQLgraph). Engineers use oInfo's text search capabilities to locate the desired concepts and properties and drag-and-drop them to construct custom nodegroups and SPARQL queries to access historical turbine test data. The TEDS use case often calls for queries involving 8+ hops between items of interest in the ontology, making SemTK's pathfinding feature indispensable. TEDS utilizes SemTK's service layer to programmatically access custom datasets to enable analytics and visualizations. By allowing engineers to develop custom queries, TEDS and SemTK have reduced the amount of time required for data access from days to minutes.

**Concept Lineage Tool.** The Concept Lineage Tool (Colt) [4], built for GE Capital, allows complex metadata to be captured on dataflows between business-critical data stores, systems and applications across Capital's global IT infrastructure, for oversight and reporting purposes. An ontology was developed to capture the notion of a data flow between systems and metadata to describe the associated data; the actual flows were represented as instances of this ontology. Colt's application developers leveraged SPARQLgraph to explore the ontology and relied upon it's drag-and-drop query generation capabilities extensively. Colt consists of a user interface that includes a rich set of features to mine and explore the data flow network, including the ability to view the network, trace and view systems upstream and/or downstream of a single node in the network, and/or view the network filtered by terms selected from the metadata taxonomy. Application developers heavily relied on SemTK's JavaScript API to interact with the triple store to accelerate the rapid development of the user interface.

## 3.2    Availability

SemTK is an open source project and the source code is available on Github[6] under the Apache License, Version 2.0. The source includes code for the SPARQLgraph web interface as well as the code for the backend Java microservices. Documentation on building and deploying the UI and the services can be found in a README file. Additional information about the SemTK service layer can be found in the wiki page for the APIs[7]. The service layer documentation includes links to their Swagger[8] documentation pages, which application developers can leverage to test the APIs. The Swagger links also allow developers to download the API documentation as JSON. SemTK has a growing wiki[9], which currently covers information on topics such as Installation, Troubleshooting, and an end-to-end "Hello World" example[10] that includes instructions for ingesting an ontology, ingesting CSV data as triples and using the drag-and-drop functionality to generate and store SPARQL queries. SemTK will be relying on Github's issue tracker for bugs and feature requests.

Additionally, users can visit SemTK's homepage[11] to use a demo version of SPARQLgraph. Users can construct SPARQL queries and upload data into an existing pre-loaded ontology or can upload their own custom ontology, ingest CSV data and query against it. The purpose of the SemTK demo page is to allow users to experiment with the existing features and capabilities before they install it locally. At this stage, we do not expect it to be persistent long-term storage for users. The triple store associated with the demo is cleared every 24 hours.

---

[6] https://github.com/ge-semtk/semtk
[7] https://github.com/ge-semtk/semtk/wiki/Services
[8] https://swagger.io/
[9] https://github.com/ge-semtk/semtk/wiki
[10] https://github.com/ge-semtk/semtk/wiki/Hello-World-SPARQLgraph
[11] http://semtk.research.ge.com/

# 4     Related Work

Previous research efforts into simplifying the use of Semantic Web technologies have focused both in visual SPARQL querying and in data triplification.

Several tools exist that enable end users to visually generate SPARQL queries to interact with triple store data. GRUFF[12] allows users to connect to arbitrary SPARQL endpoints and facilitates the visualization of a subset of the triple store's contents. GRUFF also supports visual generation of SPARQL queries allowing search classes and predicates as well as the positioning of them on a canvas and defining constraints on relationships. Similar to SemTK, GRUFF is capable of re-ordering query clauses to optimize runtime performance. OpenLink's iSPARQL [5] is another visual query generation tool, enabling connections to arbitrary SPARQL endpoints and drag-and-drop operations—iSPARQL allows users to draw SPARQL queries as a graph, but is intended for users already familiar with the SPARQL query language and does not provide significant assistance for subject matter experts unfamiliar with Semantic Web concepts or the technology stack. OWL2Query[13], a Protégé plugin, provides a visual interface to construct a SPARQL query, however it is also aimed at expert users familiar with the Semantic Web technology stack, by explicitly invoking concepts such as the A box and T box.

NITELIGHT [6] provides a query design canvas to render SPARQL queries graphically; an ontology browser and a properties inspector panel assists users in the process of visual query construction. However, as the authors state, the tool is aimed towards users with some familiarity with SPARQL. Unlike SemTK, NITELIGHT supports only SELECT queries, does not support pathfinding to connect two arbitrary classes, nor does it include any advanced filters to add constraints on a property. Schweiger et al. developed a tool also called SPARQLGraph [7] that offers a drag-and-drop interface to draw SPARQL queries. However, their tool is highly constrained, supporting connections to a limited number of RDF sources exclusively in the biological domain. It also lacks advanced pathfinding capabilities and functions to filter and add constraints on properties. Contrary to SemTK's ontology-first approach, ViziQuer [8] attempts to infer and visualize the ontology from the instance data and uses that as a starting point for users to generate SPAQRL queries. However, as the authors note, this approach quickly runs into scalability issues. Cerans et al. [9] extend ViziQuer to allow users to visually construct queries that support aggregation. Similar to ViziQuer, LODeX [10] also first infers and visualizes the schema as a graph from the instance data via a SPARQL endpoint. While visualizing the entire ontology helps circumvent the pathfinding problem in smaller graphs, this approach does not scale well for large ontologies and graphs. In terms of SemTK's visual query generation capabilities, OptiqueVQS [11] and QueryVOWL [12] come the closest. QueryVOWL lacks any kind of pathfinding capabilities, requiring the users to connect two nodes and manually identify the relationship between them. While it allows users to search for classes and instances to add to the query, there is no simple way similar to SemTK's Ontology Info panel to browse and explore the ontology. OptiqueVQS allows users to browse the domain ontology while they construct their SPARQL queries, however, it cannot find paths between two arbitrary nodes. They can show nodes that are one hop away, thus expecting users to know a priori a path from the start class to the end class, which can be tedious and difficult in large, complex ontologies. Han et al. [13] take a slightly different approach to the problem. While they allow the users to graphically draw a SPARQL query, they don't expect the users to provide class and property URIs. Users can enter strings and their system attempts to reconcile the strings to the appropriate URIs. This approach will not always lead to accurate SPARQL queries, however, since the system depends on how accurately the strings can be resolved to the correct URIs.

---

[12] https://franz.com/agraph/gruff
[13] https://protegewiki.stanford.edu/wiki/OWL2Query

SemTK expands and improves upon these tools by providing advanced features geared not toward semantic experts, but to subject matter experts who are not well-versed in OWL or SPARQL but who wish to reap the benefits of Semantic Web technologies. Through its SPARQLgraph interface's oInfo panel, SemTK allows non-expert users to first explore their domain ontologies. With SemTK's pathfinding, users are not burdened with the task of finding a path to connect two classes, which is a fairly common step in constructing SPARQL queries. Finally, SemTK makes it very simple to add constraints on both object and data properties to filter query results. We believe a combination of these three features makes it very easy for non-expert users to construct SPARQL queries.

As with visual SPARQL querying, there has been a significant amount of prior work focused on mapping CSV, spreadsheets and relational data to RDF[14]. Users are either expected to go through the cumbersome process of manually defining and writing how their data needs to be mapped to RDF triples in custom languages [14] or standards such as R2RML [15], or the tools automatically convert data into a set of RDF triples. The latter tools typically follow the model of mapping every row as an instance with most column headers mapped as properties generating local ontology mappings. These approaches fail to reuse classes and properties from existing domain ontologies that might be associated with the data. SemTK's data triplification and ingestion process falls more in the former category. Similar to RDF Refine[15], SemTK also simplifies the process by providing a simple, intuitive, graphical drag-and-drop approach to define mappings. However, unlike RDF Refine, which allows users to define arbitrary classes, properties and links between them, SemTK uses classes and properties from an existing domain ontology already provided by the user, thus conforming the data to existing schemas. In addition, SemTK's data triplification is tightly coupled with its data ingestion process. Once a mapping is defined based on the user-provided CSV, the data within the CSV is instantly triplified and uploaded to a triple store. This tight integration between triplification and ingestion makes SemTK considerably more practical and useable. Going forward, we plan to focus on semi-automating the data triplification process using similar approaches to [16] [17].

---

[14] List of Tabular to RDF Converters: https://github.com/timrdf/csv2rdf4lod-automation/wiki/Alternative-Tabular-to-RDF-converters

[15] http://refine.deri.ie

# 5    Conclusions

We present the Semantics Toolkit, which enables Semantic Web experts and non-experts alike triplify, query and manage semantic data in a user-friendly and intuitive way, and allows application developers to rapidly build knowledge-driven applications.

We detailed how users can connect to existing triple stores to retrieve domain ontologies and explore them in the SPARQLgraph user interface. We also described how users can easily generate SPARQL queries by constructing a nodegroup via a drag-and-drop interface. The process of constructing a nodegroup is highly simplified with features such as pathfinding, which helps users connect arbitrary classes in a query. As is the case in GE Power's TEDS system, these paths are often represented as multiple clauses in a single SPARQL query. Advanced functions such as suggesting legally valid values to apply filters on both datatype and object properties further simplify the query-building process. We also described how the nodegroup can be used not only for generating different types of queries, but also for generating and storing mappings to translate CSV data into triples and ingesting them into a triple store. Finally, we detailed how users can save and re-use nodegroups via a SQL stored procedure-like capability with support for runtime constraints. We also described how application developers can leverage the SemTK APIs outside of the SPARQLgraph user interface.

There are multiple areas of future work. Enhancements to the UI will not only allow for easier manipulation of large and complex nodegroups, but will also open the door for visualizing and automatically generating more complex SPARQL features such as UNION. Although SemTK is designed to work with any SPARQL1.1-compliant triple store, work remains to ensure smooth operation with a wider range of stores beyond Virtuoso. SemTK's ontology-first approach can also be balanced and expanded with the ability to use instance data in powerful ways such as: inferring an ontology from the data, optionally displaying only the subset of the ontology that is used with the given instance data, further improving query performance by reordering clauses based on the distribution of instance data, and visual exploring instance data in 2D and 3D.

SemTK remains an ongoing development effort, and is used extensively across several high-impact projects within GE. Each project benefits from the existing SemTK features and provides new opportunities for increasing its capabilities.

## Acknowledgements

The authors acknowledge the technical contributions of Ravi Palla and program support from Steven Gustafson, Matthew C. Nielsen, Arvind Menon, Tim Healy, David Hack, Eric Pool, Parag Goradia and Ryan Oattes.